# GenQ: Automated Question Generation to Support Caregivers While Reading Stories with Children

GenQ – Support Caregivers While Reading with Children

This paper talks about developing a Question Generation system that could consider cultural aspects.

Arun Balajiee Lekshmi Narayanan

University of Pittsburgh, arl122@pitt.edu

Ligia E. Gomez

Ball State University, legomezfranco@bsu.edu

Martha Michelle Soto Fernandez

University of Pittsburgh, mas731@pitt.edu

Tri Nguyen

Arizona State University, tdnguy31@asu.edu

Chris Blais

Arizona State University, chris.blais@asu.edu

M. Adelaida Restrepo

Arizona State University, laida.restrepo@asu.edu

Arthur Glenberg

Arizona State University, arthur.glenberg@asu.edu

When caregivers ask open-ended questions to motivate dialogue with children, it facilitates the child's reading comprehension skills. Although there is scope for use of technological tools, referred here as "intelligent tutoring systems", to scaffold this process, it is currently unclear whether existing intelligent systems that generate human-language like questions is beneficial. Additionally, training data used in the development of these automated question generation systems is typically sourced without attention to demographics, but people with different cultural backgrounds may ask different questions. As a part of a broader project to design an intelligent reading support app for Latinx children, we crowd–sourced questions from Latinx caregivers and non-caregivers as well as caregivers and non-caregivers from other demographics. We examine variations in question-asking within this dataset mediated by individual, cultural, and contextual factors. We then design a system that automatically extracts templates from this data to generate open-ended questions that are representative of those asked by Latinx caregivers.

CCS CONCEPTS • Insert your first CCS term here • Insert your second CCS term here • Insert your third CCS term here

**ACM Reference Format:**
First Author's Name, Initials, and Last Name, Second Author's Name, Initials, and Last Name, and Third Author's Name, Initials, and Last Name. 2023. GenQ: Automated Question Generation Systems to Support Caregives While Reading Stories with Children: This paper talks about a Question Generation System that could consider cultural aspects.. In Woodstock '23: ACM Symposium on Neural Gaze Detection, June 03–05, 2023, Woodstock, NY. ACM, New York, NY, USA, 10 pages. NOTE: This block will be automatically generated when manuscripts are processed after acceptance.## 1 INTRODUCTION

Educational researchers [13, 18, 44, 47] have found that caregiver–child interactions while reading together (dialogic reading) develop the child's long-term engagement and reading habits, giving rise to a positive home learning environment. When reading to children, caregivers can ask questions that are more open–ended; these are evocative questions [37] that require more investment in the reading and sustain the caregiver–child dialog. Caregivers could also ask Concrete (referring to content directly presented in the text), Abstract (require the reader to make an inference), or Relational (connect the text to the reader's personal experiences) questions [19]. Whereas open–ended, Abstract, and Relational questions can spark interactions between the caregiver and the child during dialogic reading, they may not always be inclined to ask those questions [39]. Hispanic/Latinx parents, have a strong oral tradition and a variety of ways in which they interact with their children, however asking questions when reading is not always part of the reading practice, this was especially evident when Latino immigrant caregivers were prompted to read as they normally would, with no other further directions [34]. Generating these questions automatically may encourage caregivers to ask these specific types of questions, motivating conversations. We consider this to be an interesting and unsolved problem in Artificial Intelligence (AI), education, and technology.

Automated question generation involves generating questions using the source text for context and answers to generate meaningful questions in the process of solving broader problems of Question-Answering [23] and Dialog systems [22], which are typically explored by researchers who build automated technological systems as solutions to process and produce text that match the characteristics of dialogue or text produced by humans in English and other languages. Using automated question generation, it is possible to limit the number of human errors that get introduced in the process of structuring different types of questions that test students' knowledge and skills [9]. Additionally, it can significantly reduce the need for human labor and enable instructors to focus on other challenging aspects of pedagogy and assessment [3]. Some well-known practical uses of question generation are in writing support systems [28], automated assessments for reading comprehension [6], vocabulary assessments [9], and multiple-choice question assessments [1]. Recently, there has been great interest in generating questions using a set of candidate answers [36, 40] from a passage [35] or an accompanying image [30, 31].

To date, though there are systems that generate questions automatically using the answers available in the text (e.g., Concrete [36, 40] or Abstract [35] questions), very few implementations address generating open-ended or Relational questions. By increasing the diversity of questions generated automatically, it is possible to develop intelligent systems that support caregiver-child dialogic reading. It is through back-and-forth conversations, initiated by diverse questions that learning happens and cultural traditions can be passed on. Scaffolding systems such as those discussed by Troseth, Boteanu and others [8, 39] for parents during dialogic reading can be enhanced using automated question generation. In our work, we extend the functionalities of an existing iPad–based intelligent application with digital storybooks [43] to support caregiver–child dialogic reading.

In this work, we explore two research questions:

**Research Question 1:** *What types of questions do US resident Hispanic/Latinx and non-Hispanic/Latinx caregivers or non–caregivers ask while reading stories with children who are between ages 5-10?*

**Research Question 2:** *How does an automated question generation system compare with the questions asked by US resident Hispanic/Latinx caregivers in terms of open–endedness, and the number of relational questions?*

One approach to building a more inclusive system would be to collect data through surveys on crowd–sourcing platforms and understand the differences in the questions that Latinx caregivers, Latinx non–caregivers, non–Latinx caregivers and non–Latinx non–caregivers ask. The rationale behind including non-caregivers in the sample is to better contrast the success or failure of our implementation in supporting caregivers with the process of DR. In our work, we adopted just that approach; we analyzed the differences in the patterns of questions asked by participants on two crowd–sourcing platforms. These participants imagined reading a few pages we selected from digital storybooks on an iPad application (ITS) [43]. We then compared the questions generated by our question generation system (GenQ) with the questions that caregivers from different demographics asked during caregiver–child dialogic reading. This work contributes to the literature by demonstrating a system that generates open-ended questions and adapts to the types of questions that US resident Hispanic/Latinx caregivers may ask.

## 2 LITERATURE REVIEW AND CONCEPTUAL BACKGROUND

Our work is inspired by the goal of building an adaptive, scalable, and cost-effective intervention to support caregiver–child dialogic reading. Further, our goal is to build a system that generates questions capable of being integrated with adaptive technological interventions. We specifically designed digital reading experiences for US resident Hispanic/Latinx caregivers.

### 2.1 Question-Asking during Dialogic Reading & Technological Interventions

For successful dialogic reading practices, caregivers can be encouraged to use various types of prompts [14, 44], which could be question types that lead to engaging interactions between the caregiver and the child.

Concrete, Abstract and Relational Questions. There are three types of questions discussed by Schwanenflugel and colleagues [19] derived from the principles of low–level and high–level questioning as discussed by van Kleeck and colleagues [41]. They refer to the question types as the CAR model and we briefly describe them here:

(1) Concrete (C) – questions that the caregiver can ask based on the text they read with the child or the pictures in the story.

(2) Abstract (A) – questions that the caregiver can ask to help the child draw inferences from the story they read together.

(3) Relational (R) – questions that the caregiver can ask to relate the content of the story with the real-life experiences that parents and children share.

They recommended that 30% of each of the three types of these questions could be asked to optimize the quality of the dialogue between the caregiver and child surrounding the reading. Latinx/Hispanic caregivers have been observed to be involved with engaging their children in the reading task, but they leave the main aspects of learning to read from formal instructions, such as the ones received in the schools. For example, parent expectations may include that teachers are primarily responsible for having children engage with diverse types of texts or that children may ask questions to the teacher in the classroom, but not to interrupt the parent with questions when reading at home. This gap in the mismatch of the roles and expectations of the Latinx/Hispanic caregivers is closed by the training offered and discussed by van Kleeck and colleagues [41]. Demir-Lira and colleagues [13] discuss the strong correlation between the different types of parent talk and especially emphasize the importance of parent talk that goes beyond the content being read together. As recommended by van Kleeck and Demir-Lira, it is possible to say that Abstract and Relational questions go beyond the content of the story. There is no strong evidence as to the proportions of these three types of questions that caregivers ought to use, although several scholars such as Troseth and colleagues as well as Whitehurst and colleagues [39, 44] discuss how different prompts have strong effects on children's language and vocabulary development [13].

*2.1.1 Open and Closed-ended Questions*

The other known typology for parent "talk" by Sun and colleagues [37] when reading stories with their child is listed below. From these different types of prompts, we derive the ideas for "Open" and "Closed"–ended questions that we described earlier.

(1) Code Talk: Pointing to specific portions in the text and highlighting the letters.

(2) Meaning Talk: Questions that expand the meaning in the text and relate the text with the child's experiences.

(3) Evocative Talk: Use of open and closed-ended questions based on the story content. Closed-ended questions could recall the content being read together.

We adapted from the prompts Troseth and colleagues [39] and the prompts discussed here by Sun and colleagues [37], to discuss our version of open-ended and closed-ended questions as:

(1) Open–ended Questions: These are questions that caregivers ask that go beyond the content being read during dialogic reading and lead to deep, engaging conversations.

(2) Closed–ended Questions: These are questions that caregivers ask that either can be answered from the content being read together ("Code" Talk or "Meaning" talk) or have a direct answer that the caregiver may expect the child to know.

*2.1.2 Dialogic Reading Practices among Hispanic/Latinx Caregivers*

Gesell and colleagues [18] discuss the importance of parental involvement in the child's education and language skill development. Specifically, they cite examples of cases where children of immigrant parents perform better in English language proficiency and score better in math when the parents take initiatives to study with their children. However, they also cite cases of low involvement from the Latinx parents in their child's learning when compared with native-born parents. They conducted a 3-month long intervention involving the CAR model for dialogic reading and detected improvements in caregiver-child interactions during dialogic reading. Considering this finding and several studies conducted by our team [19], we thought it apt to be able to develop a question generation system that takes into consideration the specific questions asked by US resident Hispanic/Latinx caregivers.

*2.1.3 Adaptive Technological interventions for Dialogic Reading*

There are a few recent adaptive technological interventions that explore the possibilities of adopting ideas from background literature in parent–child joint reading dialogs.

Xu and Warschauer [45, 46] discuss some aspects of dialogic reading with a conversational agent, wherein the agent plays the role of the "Caregiver" with a young child. They discuss the different prompts that could be implemented with such a conversational agent using Dialogflow and a Google Mini Home device. In contrast, our goal is to implement an end-to-end system to generate questions using developments from natural language generation.

Boteanu and colleagues developed an application that also implements an interface with prompts for questions parents can ask at different points in the story on TinkRBook [10].

They use a speech–based question generation system using NLTK [29] and Latent-Semantic Indexing [12] of the audio transcripts for system training and semi–automated generation. These generated prompts do not follow any theoretical framework to differentiate the question types and are based on topics of interest from textual content and speech recorded from the caregiver-child interactions during dialogic reading. In contrast, our work grounds the implementation both for our adaptive intervention (an intelligent iPad-based application with digital storybooks) and the system for question generation on prior work for Open-ended and Relational question generation. It is to be noted that, like our work, their work uses crowdsourcing to collect examples for dialog replies. Uniquely, in our work we: (1) Use crowdsourcing to collect a set of questions of specific types that we could then use to extract templates and generate questions; (2) Focus on generating questions sensitive to US resident Hispanics/Latinxs; (3) Generate questions using examples from previous sessions in dialogic reading; (4) Our implementation uses text from our intelligent iPad-based storybook instead of the speech component [5].

Alaimi and colleagues [2] investigated the benefits of pedagogical agents that ask questions based on the Gallagher and Ascher's classification scheme [16] for convergent and divergent thinking and their benefits for children's reading comprehension skills.

Although our work similarly discusses the different types of questions generated, our interest is the ability to encourage caregivers to engage in a dialog with children. Moreover, we are specifically interested in adapting to the needs of US resident Hispanics/Latinx. Finally, the question types for divergent and convergent thinking model a classroom behavior of a student in response to question-asking as opposed to our use informal learning at home use case.

Troseth and colleagues [39] discuss an adaptive technological intervention for dialogic reading, specifically designed for low Socio-Economic Status families. The intervention includes an interactive character from a popular TV show that models prompt for the parent–child dialogic reading. From their work, however, it is not clear the extent to which the questions modelled by the system for dialogic reading are generated automatically. In contrast, our implementation uses CAR questions to model dialogic reading and we implemented them in an intelligent iPad storybook for dialogic reading. This application uses the principles of Cognitive Tutor [4] within each storybook for reading comprehension.

**2.2 Question Generation Systems & Applications in Education**

The earliest known Question Generation approaches are discussed by Heilman and Smith [20] where they use a statistical approach to over generate and filter out questions based on different signals from the source text. This is among the earliest known methods to generate questions from source text without a large training set or approaches to training. With the advancements made in deep learning and improved computational power, modern question generation systems adopt a variety of approaches. Du and Cardie [15] developed the earliest known approach to generating questions using a recurrent neural network with extensive training on the SQuAD dataset by converting a Question-Answer task to a Question task. They did so by feeding the model with inputs of answers for the question along with the source text. A question generation system can be implemented in many ways. Here, we present the different pathways undertaken by prior papers, namely, (1) Supervised Learning, (2) Semi-Supervised Learning, (3) Unsupervised Learning. Among supervised learning-based approaches, the implementation by Du & Cardie [15] presents an approach to train a question generation model by providing the answers and context sentences from the SQuAD dataset [32]. They use the idea of converting a Question-Answering task into one for question generation. Many examples to generating questions in this manner exist. Among semi-supervised learning-based approaches, Kumar and colleagues [24] discuss an approach that takes a mixed approach towards unsupervised learning with pre-trained models and a small training dataset for question generation. Among unsupervised learning-based approaches, Lewis and colleagues [25] discuss the use of machine translation techniques on Cloze templates [38] originally created as a procedure by Taylor where the answers to the questions are left as blanks to be filled. Among Templates, Rules, and Statistics based approaches, Lindberg's, Master thesis [26] discusses template-based approaches as do many of the works by Heilman and colleagues [20]. These approaches are limited to the number of templates used without possibilities of extending the model for specific uses to question generation. The positive aspect to these methods of generation is they do not need pre-training or a large training dataset. Then can also directly generate questions from the source texts as input. Among some well-known, multimodal approaches, an interesting work in the space by Buddemeyer and colleagues [48] discusses the analysis that goes with developing a culturally--sensitive question generation system, that surrounds storytelling and conversations around visual cues. Specifically, their work explores the idea of question--asking related with pictures by families from different demographics and the types of questions that can be asked based on these images. In our work, we only discuss the approaches to building culturally—sensitive text—based question generation. However, the approach discussion in this prior work [48] can be suitably extended for our purposes in the application since the text in stories are often accompanied with illustrations.

In our work, we consider adopting a semi–supervised approach to generation using templates and pre-trained transformers to generate questions. Further, unlike previous works in the space of template-based question generation, our method to extract templates is agnostic of the dataset size and can be easily extended as well as scaled to the use cases for the questions generated automatically. As we will present in our work, the use of templates in our model adds to its interpretability and sensitivity to diverse demographics. Question Generation systems have become increasingly common in education technology. Gao and colleagues [17] discuss a system for second language learners. All the current state-of-art systems for question generation, including those that consider human components of question generation [27], usually generate questions that are based on the answers in the text and contextual knowledge. Using these systems for adaptive dialogic reading interventions can only generate Concrete (closed-ended) or Abstract questions at best. For the purposes of developing educational technologies that are used in classroom-based examinations, this approach to question generation works well. In contrast, our system is designed to generate questions that are driven towards creating conversations while reading stories within informal home learning environments composed of caregivers and children. Further, for effective dialogic reading, it is necessary to use Relational questions and other such open-ended questions that relate the reader to the cultural references. Finally, we specifically extract templates in our system that are representative to Hispanic/Latinx demographics in the US.

## 3 CROWD–SOURCED SURVEY EXPERIMENT

To understand the process of collecting a representative dataset of questions, we ran a crowd–sourced survey task on two platforms: Amazon Mechanical Turk (MTurk) and Prolific. Participants from these platforms were directed to Qualtrics surveys. In the following subsections, we describe our approach to survey design, deployment, and the data collection process.

**3.1 Survey Design**

We designed the survey to include the following three key tasks:

(1) Question-Asking Task: In this task, participants were shown four pages of a digital storybook from an iPad application. They were then asked to respond with the questions that they would ask while reading the page and after reading the page. For the questions that sought their responses while reading the page, they were further asked the specific sentences of the page at which they would ask the questions. A sample Question-Asking task from the survey is shown in Figure 1.

(2) Attention Check: To judge the participants' responses better, we included an attention check composed of two simple questions from the story in the Question–Asking task. If the participants answered them incorrectly, we could conclude that they did not attend to the Question-Asking task. The responses to the attention checks were verified after the participants submitted their survey responses. We accepted or rejected the responses from the participants based on their responses to the Question-Asking task and Attention Check. We did not compensate the rejected responses and discarded them from further data analysis.

(3) Demographics Questionnaire: The final section of the survey collected the specifics of participants' ethnicity, parenthood status (caregiver vs. non-caregiver), occupation (involving reading to children vs. not involving reading to children), frequency of reading with children, and a question that relates the participants to the story. The list of all factors that participants were asked about has been presented in Table 1. Participants were required to complete the survey in 15 minutes and were compensated 2.10 USD for every successful submission.

**3.2 Survey Deployment**

To crowd–source the responses in our survey, we deployed studies on Amazon Mechanical Turk and Prolific separately. We used a variety of filters offered by these platforms to select the target demographic group. Our final sample sizes on MTurk for non-caregivers and caregivers were 30 and 19 respectively, whereas we recruited 250 participants from Prolific. These were the final sample sizes after discarding data that weren't suitable for analyses. In most cases, we could recruit distinct participants for the two surveys we deployed for two different stories on the crowd–sourcing platforms. However, we had a nearly similar number of participants who responded to the surveys on both the stories. We list details of the deployment in Table 2.

Table 1: Questions from Demographics Section on our Survey

| Question | Possible Responses |
| --- | --- |
| Are you a caregiver? | Yes/No |
| Does your occupation involve reading to children? | Yes/No |
| How many times have you read to young children in your life? | Rarely (0-4) times |
|  | Sometimes (5-24 times) |
|  | Frequently (25-99 times) |
|  | Very Frequently (>100 times) |
| This past month, how much time have you spent reading to a child? | None at all |
|  | 0-30 minutes |
|  | 30-60 minutes |

| Question | Possible Responses |
| --- | --- |
| When you read to children regularly, about how many times in a week would you do so? | 1-2 hours<br>More than 2 hours<br>I never regularly read to children<br>1-2 times<br>3-5 times<br>More than 5 times |
| How do you find the experience of reading to children? | Very Unenjoyable<br>Somewhat Unenjoyable<br>Neither enjoyable nor unenjoyable<br>Somewhat enjoyable<br>Very enjoyable |
| Question that relates to the story content | Yes/No |
| Are you Hispanic/Latinx? | Yes/No |
| Which of the following most accurately describes you? | List of Race per U.S Census |

*3.2.1 Adaptive Technological interventions for Dialogic Reading*

We chose two distinct stories from an iPad application for digital storybooks for the question-asking task on the survey. These stories had slight differences in terms of how they were contextually and culturally situated. Whereas one story was on the life and activities of people in rural America as shown in Figure 1, the other touched on the aspects of Latinx/Hispanic culture as shown in Figure 2. These stories had key elements as well as illustrations to them that depicted the typical aspects related to the two diverse cultures. The question-asking tasks in the survey for the two stories were identical in that participants had to respond with the questions they would ask while reading stories with children. We presented these prompts during reading the page and after reading the page. We were motivated to design our question prompts in this way because [37] noted differences in caregiver talk before, during and after reading stories with children.

We customized some questions in the attention checks and demographics questionnaire based on the content of the story. For the story about rural America, we asked questions such as "Which animal did not appear in the story?" for the attention check question. We expected that participants who may be familiar with the situations discussed in the story may relate to the stories better. Hence, in the demographics section, we asked "Have you been to a farm before?". For the story on Latinx/Hispanic culture, the attention check questions were "How many pesos did the mother give Sofia?" and a question in the demographics section asking "Have you been in a celebration in the style of this story?". The choice of these questions was motivated by our expectation that participants familiar with the situations in the story may relate better.

We focused on recruiting participants by "residence" in the US as opposed to "citizenship" in the US to be more inclusive and mindful of the fact that our target demographics (Hispanics/Latinxs) may not necessarily be US citizens.

*3.2.2 Deployment on MTurk and Prolific.* MTurk offers a variety of filters for the Requesters on the platform to select their participant pool. In our case, we chose the "Parenthood Status" filter to select whether participants were caregivers. Prolific offers a variety of filters for Researchers on the platform to selectively choose the participant pool. In our case, we chose filters for participants who are:

(1) Bilingual (speak one native language and English/one other language), Latinx/Hispanic and Caregivers
(2) Bilingual (speak one native language and English/one other language) and Latinx/Hispanic
(3) Caregivers and not Latinx/Hispanic

(4) Not Caregivers and not Latinx/Hispanic

*3.2.3 Qualitative Coding: CAR & Open/Closed-Ended Questions*

After completing data collection on the crowd–sourcing platforms, we coded the responses for further analysis. As a step towards preprocessing, for entries where the participants entered multiple questions that they could ask on a given page, we split the rows until the number of questions per entry were one per page for during reading the page and after reading the page. We followed the same coding procedure to qualitatively code the responses for Concrete (C), Abstract (A), and Relational (R) questions as well as Open(O)/Closed(C) ended questions. Two coders from the research team coded up to 50% of the dataset collected on the Amazon Mechanical Turk (MTurk) followed by independent coding of the remaining dataset collected on Prolific. For the two coding tasks, the interrater reliability was Cohen's Kappa $\kappa = 0.92$ when coding questions as Concrete, Abstract, or Relational. When checking agreement for the Open/Closed-ended questions coding task, Cohen's Kappa $\kappa = 0.94$. We did not find characteristic differences for all the responses collected on the two platforms for crowdsourcing; hence, we think that our code book for Concrete, Relational, and Abstract and Open/Closed-ended questions was sufficiently robust to code all responses collected from both the platforms.

## 4 CROWDSOURCED SURVEY RESPONSE ANALYSIS

Before proceeding the discussion on the survey data analysis, we begin by revisiting the research question addressed in this section:

RQ1: What types of Questions do US resident Hispanic/Latinx and non-Hispanic/Latinx caregivers and non-caregivers ask while reading stories with children who are between ages 5-10?

From the collected dataset, we found that the number of Relational questions asked by participants on Prolific overall (N=250) lies between 0 and 7 (mean=1.47,sd=1.82) and the number of Abstract questions asked by the participants lies between 0 and 8 (mean=3.82,sd=2.25). We also found that the number of open-ended questions asked by participants on Prolific overall (N=250) lies between 0 and 8 (mean=5.22,sd=2.59). This includes the number of Relational questions asked by participants of both surveys and of one of two surveys. We present the distribution of the number of Abstract, Relational and Open-ended questions asked by participants from different demographics on the two surveys in Figures 3 and 4. To handle the cases where participants responded to both surveys on the two stories, we split the dataset into participants who responded to one survey (N=208, number of Abstract mean=3.76, sd=2.24; Relational questions mean=1.53, sd=1.88; open-ended questions mean=5.28, sd=2.66) and who responded to two surveys of the two stories (N=42, number of Abstract mean=4.07, sd=2.28; Relational questions mean=1.16, sd=1.44; open-ended questions mean=5.06, sd=2.3). We notice that the number of Relational questions or the number of open-ended questions does not follow a normal distribution and the numbers in the Latinx, Latinx caregivers, non- Latinx caregivers and non-Latinx non-caregivers do not have equal variances. Further, considering that our data was right-skewed, and we have count-based data, we conducted negative binomial regression tests. We proceeded with testing for significance with negative binomial regression on the responses from the participants (using MASS [42] package, R). First, on the subset of responses from participants of one survey, we attempted to discern if there are any differences in the number of Abstract, Relational, and Open-ended questions between the two stories used on the surveys. We found that there is a significant difference in the number of Relational questions asked by participants for the story "Celebrations" (mean=0.7126, se=0.121, p=0.00025) when compared with the story "Best Farm" (mean=0.0461,se=0.136). We also found a similar result in terms of the number of Open-ended questions asked on "Celebrations" (mean=1.76, se=0.0507, p=0.00969) when compared with "Best Farm" (mean=1.56, se=0.0546). We did not notice any significant effects because of the story on the survey on the number of abstract questions. We present more comparisons among different factors in Table 3. For the same subset, we then examined if the number of Abstract, Relational, and Open-ended questions asked on the responses depend on the participants being Latinx or non-Latinx and on the story they respond to. We did not observe any significant interaction effects by the demographics in terms of the two stories in the number of Relational, Abstract, or Open-ended questions.

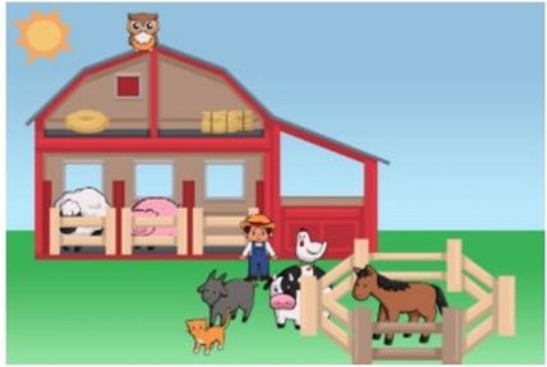

**Figure 1:** Our survey shows four similar pages as shown from a story of a farmer trying to win a prize for the "Best Farm." The main task in total involves eight forms for questions that participants would ask while reading the page and after reading the page.

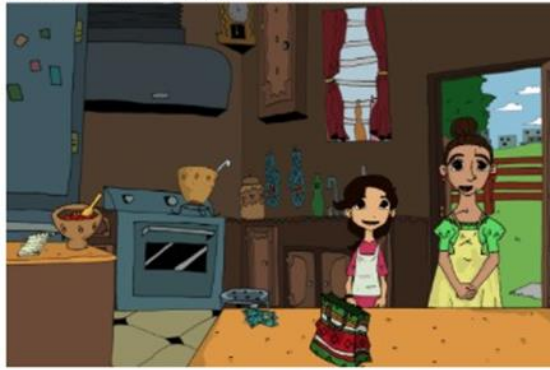

**Figure 2:** Our survey shows four pages like the one shown from a story of a young girl named Sofia helping her mother cook traditional dishes such as mole and champurrado for her sister's wedding

Table 2: CrowdSourced Survey Recruitment Summary (*Human Intelligence Tasks)

| Platform | Criterion | Description | Sample Size for the Study |
|---|---|---|---|
| Amazon Mechanical Turk | Age | over 18 years | |
| | Residence | United States | |
| | Approval Rate | >= 90-95% | |
| | # of Approved HITs* | >= 100-1000 | |
| | Demographics | Caregivers | 30 (caregivers) |

| Platform | Criterion | Description | Sample Size for the Study |
| --- | --- | --- | --- |
| | | Non-Caregivers | 19 (non-caregivers) |
| Prolific | Age | | |
| | Residence | over 18 years | |
| | Approval Rate | United States | |
| | Bilingual | >= 90% | |
| | Demographics | One native language + One other language | |
| | | Non-Latinx/Hispanic Non-Caregivers | 43 |
| | | Latinx/Hispanic Caregivers | 43 |
| | | Latinx/Hispanic Non-Caregivers | 58 |
| | | Non-Latinx/Hispanic Caregivers | 48 |

Further, we examined whether the number of Abstract, Relational, and Open-ended questions asked on the responses depend on the participants being caregivers or non-caregivers and on the story they respond to. Again, we did not notice any significant interaction effects by the demographics in terms of the two stories in the number of Abstract, Relational, or Open-ended questions.

Finally, we examined whether the alignment between the responses to Relational questions ("Have you been to a farm before?" or "Have you been in a celebration in the style of the story?") and the content of the story affects how many Abstract, Relational, and Open-ended questions participants ask on the survey. We detected a significant interaction effect between the past experiences of the participants that are related to the story. This was stronger for the story "Celebrations" (mean=1.32, se=0.0614, p=0.0276) than "Best Farm" (mean=1.28, se=0.0961). We present the results of the tests (non-standardized coefficient estimates) in Table 4.

For the repeated participants, we contrasted the number of Relational and Open-ended questions on "Celebrations" (CB) with "Best Farm" (BF). Because few participants completed the survey twice and the distribution is non-normal, we performed the Wilcoxon Rank Sum test (nonparametric equivalent of the two-sample t-test) . We found the number of Relational questions to be greatest when the participants asked more questions in the first story ("Celebrations") than the second story ("Best Farm"). See Table 5.

Table 3: Overall Frequencies in Reading stories with children by Demographics and the differences in the mean number of Relational questions

| Caregiver | Hispanic/Latinx | Mean Frequency (SD) | Mean # Relational (SD) |
| --- | --- | --- | --- |
| Yes | Yes | 2.85 (1.44) | 1.18 (1.86) |
| No | Yes | 2.27 (1.02) | 1.70 (1.95) |
| Yes | No | 3.16 (1.28) | 1.44 (1.72) |
| No | No | 2.58 (0.92) | 1.64 (2.04) |

Table 4: Results from the Negative Binomial Regressions on the number of Abstract, Relational and Open-ended questions in the responses from participants of one survey (***p<0.0001, **p < 0.01, *p<0.05, p<0.1)

| Factors | Outcome (# questions) | Coefficient Estimate | Coefficient Std. Error | Z-value | AIC | 2xloglikelihood | p-val |
|---|---|---|---|---|---|---|---|
| Story | #Relational | 0.66647 | 0.18201 | 3.662 | 669.47 | -693.475 | 0.0002 |
| Story & Latinx | #Relational | 0.1834 | 0.3684 | 0.498 | 703.23 | -693.225 | 0.61 |
| Story & caregivers | #Relational | 0.07669 | 0.36487 | 0.210 | 701.88 | -691.877 | 0.833 |
| Story & Real-Life Experience | #Relational | 0.1410 | 0.4790 | 0.294 | 702.72 | -692.717 | 0.76 |
| Story | #Abstract | -0.0702 | 0.08472 | -0.933 | 926.01 | -920.014 | 0.35 |
| Story & Latinx | #Abstract | -0.14018 | 0.17192 | -0.815 | 927.09 | -917.088 | 0.41 |
| Story & caregivers | #Abstract | 0.07425 | 0.16865 | 0.440 | 927.99 | -917.99 | 0.66 |
| Story & Real-Life Experience | #Abstract | -0.5023 | 0.2281 | -2.202 | 923.23 | -913.23 | 0.027 |
| Story | #Openended | 0.19269 | 0.07449 | 2.587 | 1010.7 | -1004.704 | 0.0096 |
| Story & Latinx | #Openended | -0.11716 | 0.15095 | -0.776 | 1011.3 | -1001.30 | 0.43 |
| Story & caregivers | #Openended | -0.07393 | 0.14904 | -0.496 | 1014.2 | -1004.242 | 0.61 |
| Story & Real-Life Experience | #Openended | -0.3241 | 0.2167 | -1.496 | 1011.7 | -1001.686 | 0.13 |

## 4.1 Comparisons with CrowdSourced Dataset and Human Annotations

While it is possible to analyze our crowd–sourced dataset in isolation, we could draw more insights by comparing the dataset with other datasets that were collected by fellow team members on the same project using crowd–sourcing and identify if the differences in the strategy of data collection affects the data that is collected. Further, we compared our dataset with a small sample set of questions generated by the research team members for the two stories – The Best Farm ("Best Farm") and A Celebration to Remember ("Celebrations") and notice if there are other differences between the types of questions that the research team could come up with in comparison with caregivers and non-caregivers from the U.S. resident Hispanic/Latinx or non-Hispanic/Latinx communities on a crowd–sourcing platform.

To perform this comparison, we coded another dataset collected independently by the team members on Amazon Mechanical Turk (MTurk) with a slightly varied survey design. In this survey, the team members asked two sets of questions on the main task, which we paraphrase here,

(1) Imagine reading the story with a child and type questions that you would ask them while reading the specific page shown here, in the field below. Indicate the sentence at which this question could be ask.

(2) Imagine reading the story with a child and type questions that you would ask them while reading the specific page shown here, in the field below. Indicate if the question typed in the field is Concrete (C), Abstract (A) or Relational (R).

The two survey questions were repeated for all the pages of all the chapters for different stories on the digital storybook in an iPad application described earlier. The pages from the application had the picture accompanying the text from the story verbatim. Around 19 responses were collected for The Best Farm for all pages of the story and another 19 for A Celebration to Remember.

From the data collected, we sub-sampled only these questions responded to by the participants of the survey, for the first four pages that match with our survey design. A researcher from the team independently coded the dataset for Concrete (C), Abstract (A) and Relational (R) questions, using the codebook developed and described earlier. We then performed an overall comparison of the proportion of Relational questions asked for the four pages by each participant of this survey with that of proportion of Relational questions asked for the four pages by each participant in the dataset we collected with our survey for the same two stories. We also compared the proportion of Relational questions asked overall by the participants to our survey with the proportion of Relational questions in the small sample set created by our research team members for the same two stories. We present these results in Table 6.

### 4.2 Word Usage Analysis

Apart from the analysis on the dataset in terms of differences in the number of Relational questions, we observed the differences in the types of questions by demographics using automated natural language analysis to qualitative compare the differences that are not captured using the CAR or Open/Closed ended coding scheme. From the dataset, we separated all the Relational questions asked by participants from different demographics. As suggested by the statistical tests, we notice that there are relational questions asked by participants from all demographics for both the stories in equal measure.

Table 6: Mean length of Relational questions in the overall survey responses

| Demographic Group | MTurk (Previous) | Study Team | MTurk/Prolific (Current) |
|---|---|---|---|
| Latinx Parents | NA | NA | 9.54 |
| Latinx non-Parents | NA | NA | 9.17 |
| non-Latinx Parents | 11.07 | 11.33 | 9.64 |
| non-Latinx non-Parents | NA | NA | 9.29 |

[a] This is example of table footnote.

To further understand the demographics related differences, we ranked the top-10 relational questions asked by participants from each demographic and analyzed the word usage patterns. Our findings:

(1)     Finding 1: Based on the similarities of the different questions, it is impossible to capture the differences between demographics on a crowdsourcing platform.

(2)     Finding 2: Caregivers may or may not always refer to themselves when posing the questions about the story.

(3)     Finding 3: Most of the relational questions are posed by asking the child to relate to the situation and thinking about how they would react. Further, these questions also sometimes make the child think of actions in relation with their parents/caregivers.

(4)     Finding 4: Caregivers ask questions that either be asking them about their affinity to certain characters or situations in the story while others ask if they have been in the situation of the story before or what they would do if they were in the situation like the ones in the story.

From these findings, it becomes clear that Finding 4, if extended could possibly become more culturally influenced based on the traits children learn from their parents. Further, these questions could be posed by parents with a neutral stance rather than a personally mutually way so that the child could possibly become more acquainted with different aspects of the culture as well as familial traits. Also, the child could get to know more about their culture if these questions are slightly modified by the parent when posing these questions to start a conversation. The modifications and the differences by demographics or specific culture are difficult to be captured using a crowdsourced dataset but more possible in a co-design or interviewing the families from different cultures in semi-naturalistic environments and experimental setups.

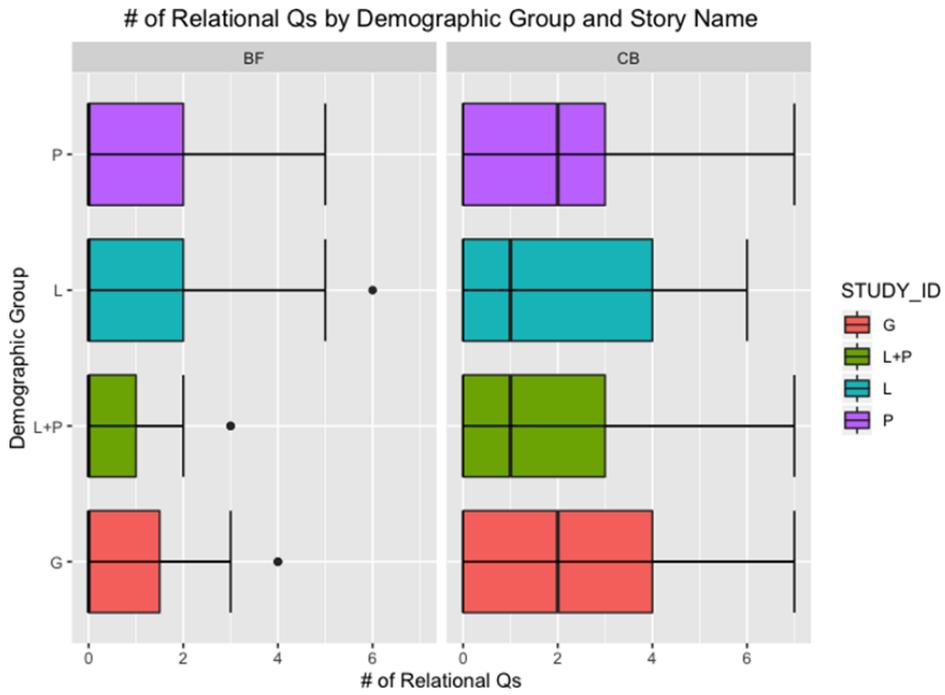

Figure 3: Total number of Relational questions asked by participants on the two stories "Best Farm" (BF) and "Celebrations" (CB) (G="Non-Hispanic/Latinx Non-Caregivers", L="Latinx Non-Caregivers", P="Non-Hispanic/Latinx Caregivers", L+P="Hispanic/Latinx Caregivers")

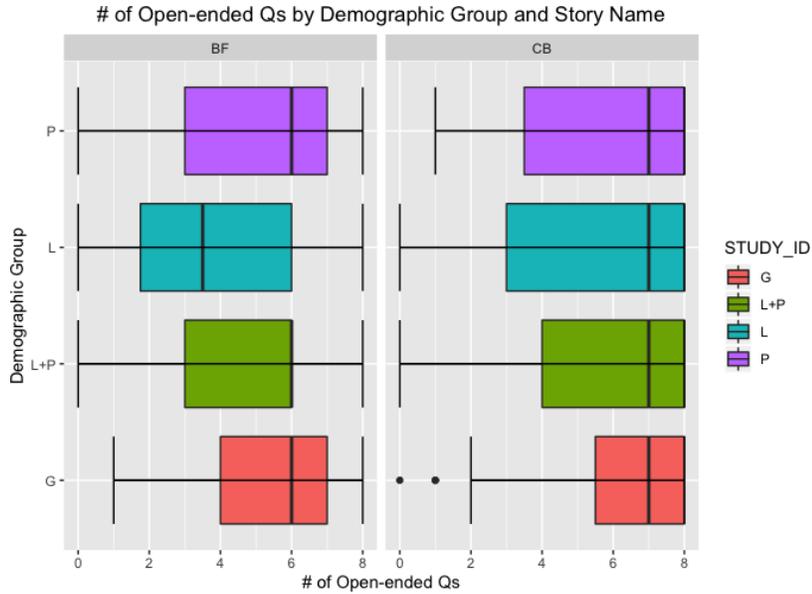

Figure 4: Total number of Open-ended questions asked by participants on the two stories "Best Farm" (BF) and "Celebrations" (CB) (G="Non-Hispanic/Latinx Non-Caregivers", L="Latinx Non-Caregivers", P="Non-Hispanic/Latinx Caregivers", L+P="Hispanic/Latinx Caregivers")

Table 5: Results from the Wilcoxon Rank-Sum test on the number of Relational and Open-ended questions in the responses from participants to both the stories on two separate surveys (***p<0.0001, **p < 0.01, *p<0.05, .p<0.1)

| Grouping Variable | Outcome | W-value | p-value |
| --- | --- | --- | --- |
| Story | #Relational | 111.5 | 0.004118 * |
| Story & Latinx | #Relational | 48.5 | 0.0717 |
| Story & caregivers | #Relational | 154.5 | 0.2435 |
| Story & Real-Life Experience | #Relational | 166.5 | 0.915 |
| | | | 0.372 |
| Story | #Abstract | 183 | 0.6663 |
| Story & Latinx | #Abstract | 81 | 0.9142 |
| Story & caregivers | #Abstract | 200.5 | 0.8511 |
| Story & Real-Life Experience | #Abstract | 163.5 | |
| Story | #Openended | 190 | 0.1025 |
| Story & Latinx | #Openended | 116 | 0.5366 |
| Story & caregivers | #Openended | 160.5 | 0.1286 |
| Story & Real-Life Experience | #Openended | 203 | 0.621 |

## 5 AUTOMATED QUESTION GENERATION WITH TEMPLATES

Our two–step implementation follows the methodology discussed in prior template-based approaches to question generation [20, 26]. Our model (a) extracts viable templates from the source dataset of questions and (b) then generates open-ended questions that caregivers could ask while reading stories with children – Concrete (C), Abstract (A), Relational (R). In the following subsections, we discuss these two steps in greater detail.

### 5.1 Base Template Extraction

The first step towards generating questions without sufficient training data involves template extraction. This means the process of constructing useful templates for generating open-ended and CAR questions. We perform this using the following two sub steps:

(1) Collate the small sample seed dataset
(2) Convert the questions into generic templates for the question types.

We utilized the data we collected on MTurk (777 questions for 3 different stories featured on an iPad digital storybook) by team members from the larger project of the studies discussed in this paper. We pre–processed and cleaned this dataset. Next, we converted these questions into Part-of-Speech (PoS) Tag sequences using SpaCy [21]. We categorized these templates into open and closed–ended questions or CAR questions depending on the types of questions that were to be generated. All steps in the process are fully automated.

## 5.2 GenQ: Base Templates to Generate Questions from Story Texts

Our system for question generation selected the most relevant templates by matching the PoS tag sequence of a source sentence and question templates. For instance, for a sentence shown in Table 7, The ground shakes and slides...the buildings move. It's an earthquake! (the punctuation is excluded), the template for a Concrete question was matched {"What","AUX","NSUB"}, that is, the PoS tags of the words in the sentence have corresponding tags in the template.

Our GenQ replaced the words with their corresponding PoS tags in the matched template. Here, we replaced the words in the sentence with PoS tags "AUX" and "NOUN" to generate What is earthquake? Finally, we used pretrained transformers [11] for paraphrasing tasks to correct the question grammar and denoise the generated questions. In this example, the question is converted into What is an earthquake? Our GenQ generates questions for all sentences by page. All three steps above were fully automated from the preprocessing of the input data to the generation of questions. Based on the type of the input data, the system uses different templates to generate the question.

**Table 7:** GenQ generated question compared with questions by human.

| Question Type | Source Text | Human QG(MTurk) | GenQ |
|---|---|---|---|
| Concrete (C) | The ground shakes and slides...the buildings move...It's an earthquake! | What causes an earthquake? | What is an earthquake? |
| Abstract (A) | It was important to leave a hole at the top of the teepee | Why did they leave a hole at the top of the teepee? | Why was a hole left up? |
| Relational (R) | "You can buy whatever you want with the change" she said to Sofia.. | What would you buy with the change left over? | What can you buy with the change? |

We believe that the steps discussed above make our system more flexible and interpretable. By utilizing and constructing templates from the sample seed dataset, we can easily observe the differences created by subtle variations in the arrangement and order of different phrasings of the questions. This would ultimately lead to our system being more sensitive to the questions that participants submit on the survey. Some examples of the CAR questions generated by the system are presented in Table 7. We also present a simplified depiction of the question generation pipeline of our model in Figure 5.

*5.2.1 Open/Closed-Ended Question Template Extraction and Question Generation*

Our procedure to generate open and closed–ended type questions involved the process of data augmentation. Along with the training set of questions used for the base template extraction, we append the templates for Open/Closed–ended questions. To do this, we use a separate set of responses collected on Prolific for the two stories from the surveys, as shown in Figures 1 and 2. Overall, we collected 337 questions for 4 pages (during and after reading) for the two stories.

We extracted the templates for open and closed ended questions from these responses to the survey using the mechanism discussed in the subsection 5.1 but instead of the Concrete/ Abstract/ Relational coding scheme, we utilized our open/closed-ended coding scheme. That is, we extracted, categorized and saved the templates for the questions coded as open-ended (O) and closed-ended (C) questions to assist the question generation step of the pipeline. We then generated the mechanism using the template-based generation as discussed before:

(1) Template-filling: We fill the extracted templates with the appropriate words from the source text that match the tags

(2) Paraphrasing: We then paraphrase using pre–trained transformers [11] the template- based generated questions to correct grammar and make the questions more semantically meaningful.

## 6 REPRESENTATIVE TEMPLATES FOR QUESTION GENERATION

Before discussing the qualitative comparisons between the framework to generate questions and the questions asked by participants of the survey in the US, we revisit research question 2:

RQ2: How does an automated question generation system compare with the questions asked by Hispanic/Latinx caregivers in the US in terms of quality, open-endedness, and the number of relational questions?

From the collected dataset, we split the questions asked by Latinx, Latinx caregivers, non- Latinx caregivers and non-Latinx non-caregivers. We proceeded with the template extraction on these questions and performed a TF-IDF [33] score computation for all the words and tags used in the extracted templates. We then calculated the scores for each template by adding up the 2-norm of each word's individual column scores and summed those values. The final rank (top 50 or top 100) was computed using these scores in descending order. We present the results and comparisons in Table 8. We also present sample templates that were extracted for the different demographics in Table 9. Using the proof of concept from Table 7, we can say that if these templates are being selected in the process of extraction, then they will necessarily be used for the generation of questions. That would mean that we are able to, using a representative small seed dataset, incorporate a system that can generate questions for diverse demographics. Most of the questions generated by our system needed a second stage of human editing, hence, while not being perfect, this system is able to generate questions that are of the open–ended or relational question type.

## 7 DISCUSSION

Reading provides a wealth of opportunities for parents to interact with their children. One way of generating this interaction is by asking questions. Hispanic/Latinx parents, have a strong oral tradition and a variety of ways in which they interact with their children. Asking questions to their children during reading may become part of their default reading practices. This is especially true after participating in learning sessions as parents were more prone to asking a greater number of questions during reading [19]. The great benefit of our Question Generation system (GenQ) is that it can scaffold parents with immediate feedback via an automated system that provides parents with a variety of types of questions in real time. GenQ may make the parents more aware of the myriad possibilities regarding question types and the moment during reading when those questions are more suitable to ask. Having a conversation generated by the back-and-forth questioning may also be an avenue for the dyad to jointly enjoy reading.

One of the greatest contributions of our system is that it takes into consideration open-ended questions, making this GenQ more culturally relevant. The nature of the open-ended questions is to have the parents think of something related to their personal experience and relate it back to the story. Given the richness of oral tradition in the Latinx culture, it may come more natural for parents to incorporate questions of this nature. Using the questions generated by our system, parents can extend their conversations with their children not only to events related to their lives but also to other types of conversations that are culturally relevant. We understand that questions are just one strategy that parents can use to foster meaningful conversations, thus our system considers questions as the springboard for conversations that can foster deeper understanding of specific topics or questions that can provide an opportunity to explore traditions and values that may otherwise go unexplored without a starting question. The conversations that may follow-up are dependent on the context and the unique experiences of the family—which we hope spark with the questions automatically generated by the system.

When designing our question generation system, we endeavored to have human input and coupled that with the automation of the system. Our unexpected result was not observing any significant differences in the number of Relational questions when compared between Latinx and non-Latinx as well as between caregivers and non-caregivers, in future iterations, we will take this into consideration when designing our system. The other aspect that needs further exploration as to what could the other contributing factors be that lead to the significance in the difference between the stories that we chose for the two surveys. Whereas the story "A Celebration to Remember" ("Celebrations") indeed is representative of the Latinx culture, but more specifically of a Mexican-American tradition, it may be possible that the participants on crowd–sourcing platforms are not a representative enough pool of all the US resident Latinxs and the diverse within Latinx cultural backgrounds they come from. This potential limitation of the sampling procedure possibly not being as representative as the target population could hinder the development of representative & sensitive AI systems. Hence, it is possible that we may have noticed significant differences in our results if the samples were more representative that they are already. Perhaps this could be addressed by having a representation of the different cultural differences within the Latinx population.

Importantly, because we have human inputs as a base for comparison with the questions generated by our system (discussed in Table 7), we are on the right track towards generating a system that provides questions that are sensitive to our target population.

Further, it is possible to observe differences in the during and after page responses in terms of the number of Relational and open-ended questions. However, given the limitations on the data collected through crowdsourcing, it seems to be that this analysis may require us to consider more authentic ways to collect data to allow for robust data analysis. Another limitation of using crowdsourcing is the potential for the same participants to complete the survey more than once. This limitation is important to address in future research and something that researchers using it for data collection should take into account.

Even though we could generate some human-like questions, most of the questions generated by our system needed a second stage of human editing. This could be addressed as future work in the space of human–centered AI systems. An application could be developed to allow human volunteers to revise machine-generated questions and perfect them for practical use. These finalized questions could then be rated for engagement, clarity, and educational potential by parents. Our template extraction eliminates the possibility of misrepresenting the data. If the dataset that is used to train and implement a question generation system using questions templates that are generative, it should necessarily produce representative questions for different target demographics.

Some of the limitations that we noticed in our work could be addressed as upcoming research directions. First, we limit our template extraction to support for a handful of Part- of-Speech tags (PoS tags) for simplicity of the generation. However, to build a more robust system, it could be possible to include support for more Part-of-Speech tags. We also limit our implementation slightly with the use of a pretrained transformer, which could add its own noise and demographic bias, making our overall system less representative. We are currently exploring more approaches to correct grammar that does not involve a less interpretable system yet achieve the same efficiency. Further, relational questions may sound absurd if the child has not had proper experience with that situation, such as asking about a trip to the zoo when the child has never been to a zoo. Additionally, providing questions to the parents is likely less effective in teaching the parents how to generate questions than having the parents generate their own questions (and providing feedback) [7].

**Table 8:** Extracted Template Sensitivity

| Rank | Template Demographics | Template Proportions |
|---|---|---|
| Top 50 | Parents/Latinx/Latinx Caregivers/Non-Latinx non-Caregivers | 30%/24%/22%/24% |
| Top 100 | Parents/Latinx/Latinx Caregivers/Non-Latinx non-Caregivers | 25%/20%/31%/24% |

**Table 9:** Extracted Template Demographics

| Demographic | Top-1 Templates Extracted |
|---|---|
| Latinx | 'have your NSUBJ ever forgotten important DOBJ PREP DET POBJ' |
| Caregivers | "do NSUBJ ROOT why Sofia's horse is named Mancha?" |
| Latinx Caregivers | 'What AUX NSUBJ ROOT NSUBJ and her mother AUX do AUX fin DET missing DOBJ' |
| non-Latinx non-Caregivers | 'Have your NSUBJ ever ROOT cooking only AUX find out that NSUBJ forgot AUX buy DET DOBJ' |

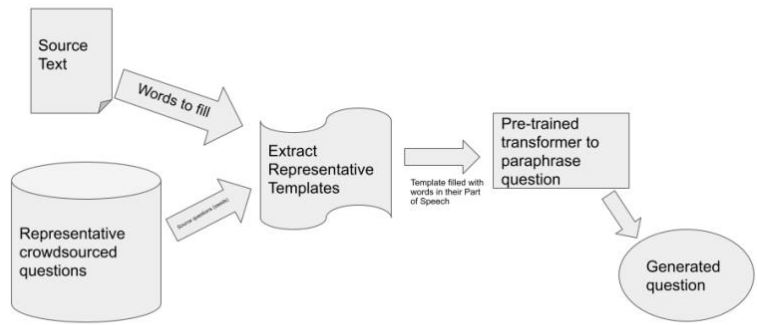

Figure 5: A simplified diagram of data flow to generate questions from source text. The system uses words from the source text alone along with a set of extracted templates from a seed dataset of questions to generate questions. The questions are paraphrased using a pre-trained transformer to generate a grammatically correct and semantically meaningful questions.

Overall Practical Implications. The work presented here has 3 main implications for educational settings First, Adopting the system in a classroom or lifelong—learning context: A system like the one we are building has the potential to assist instructors with the automated question generation to build questionnaires for assessments, it can also be used to build a dialogue between the teacher and the student in assisting teachers to ask open-ended questions to make students in their class more engaged in the course material. Second, the GenQ can assist teachers in adopting strategies for question-asking as a part of dialogic reading towards a common generalizable framework. 3) Having these questions as conversation starters that can be used in the home. Given the availability of technology in the home, it is possible to imagine that families can adopt the system as part of their reading practice. Potentially, teachers could bridge home-school connection by providing questions generated by the system to the caregiver about a specific topic for further exploration based on their own experiences. This would be an important support for all parents, but especially useful for parents who are not familiar with the content, or bilingual households who can benefit from having questions generated in both in English and Spanish.

Despite some of these merits we discussed above, our work has as some limitations. Firstly, while we consider some aspects from the source text of reading comprehension, we didn't dig deeper into the demographic variables from our crowdsourced platform to know the participant educational background and their skills of abstraction. This could indeed play a role in the types of the questions they generate, which we plan to extend as future work. We also think that changing the age range of text to suit better for readers from 5 years old to 10 years old may also change the generated questions. Finally, another interesting direction to take this work further would be to use ChatGPT in combination with culturally responsive templates to finetune its generation process or use ChatGPT to generate certain kinds of templates first [49, 50] which then finetune the generated questions to be more culturally responses.

## 8 CONCLUSION

In this work, we presented our explorations on the diversity and nature of questions asked by caregivers and non-caregivers from Hispanic/Latinx or non-Hispanic/Latinx communities in the US. We also present our extensions to the extracting templates to generate questions automatically that are sensitive to the Hispanic/Latinx demographics. From our work, we intend to show the possibilities of building systems that are adaptive to the Hispanic/Latinx by constructing datasets that are representative of the questions they ask. This way we believe our work could lead to further research and possibilities to building adaptive technologies that are sensitive to different demographics with inclusive design and development.

## ACKNOWLEDGMENTS

This work was supported by the National Science Foundation Award Nos. CISE-IIS-1917625 and CISE-IIS-1917636. We thank Dr. Erin Walker extensively for their thoughtful suggestions, edits and help with all the ideas discussed in this work. We also thank Dr. Malihe Alikhani and Varun Gangal for their ideas. We thank Sarah M. Fialko for her contributions to the project implementation. We also acknowledge the discussions with different members from our lab (Dr. Perez-Cortez, Jay Patel, Amanda Buddemeyer., Dr. Lobczowski) and the extended research team with helping the authors in fine-tuning the ideas presented in this work as well as with proof-reading/corrections to the text.